\pdfoutput=1

\documentclass[11pt]{article}

\usepackage[]{EMNLP2023}

\usepackage{times}
\usepackage{latexsym}
\usepackage{comment}
\usepackage{url}
\usepackage{graphicx}       
\usepackage{amsmath}
\usepackage{amssymb}
\usepackage{amsthm}
\usepackage{mathtools}
\usepackage{booktabs}

\usepackage[T1]{fontenc}

\usepackage[utf8]{inputenc}

\usepackage{microtype}

\usepackage{inconsolata}

\usepackage{hyperref}

\newcommand{\eg}{\emph{e.g.,} }
\newcommand{\ie}{\emph{ie.} }
\newcommand{\sota}{\emph{state-of-the-art }}

\DeclareRobustCommand{\ourlogo}{%
  \begingroup\normalfont
  \vspace{-0.2em}%
  \raisebox{-0.4em}{%
  \includegraphics[height=1.5em]{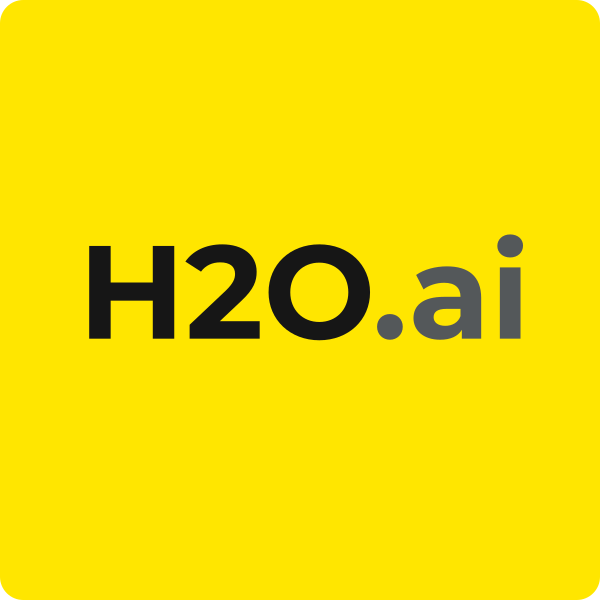}%
  }%
  \kern 0.4em%
  \endgroup
}

\title{\ourlogo H2O Open Ecosystem for State-of-the-art Large Language Models}

\author{
Arno Candel, 
Jon McKinney, 
Philipp Singer, 
Pascal Pfeiffer, \\ 
{\bf Maximilian Jeblick}, {\bf Chun Ming Lee}, {\bf Marcos V. Conde} \\
\\
{\bf H2O.ai, Inc.}\\
Mountain View, CA\\
\texttt{\{firstname.surname\}@h2o.ai} \\
\url{https://gpt.h2o.ai/} \\ 
}

\begin{document}
\maketitle


\begin{abstract}
Large Language Models (LLMs) represent a revolution in AI. However, they also pose many significant risks, such as the presence of biased, private, copyrighted or harmful text. For this reason we need open, transparent and safe solutions.
We introduce a complete open-source ecosystem for developing and testing LLMs. The goal of this project is to boost open alternatives to closed-source approaches. We release h2oGPT, a family of fine-tuned LLMs of diverse sizes. 
We also introduce H2O LLM Studio, a framework and no-code GUI designed for efficient fine-tuning, evaluation, and deployment of LLMs using the most recent state-of-the-art techniques.
Our code and models are fully open-source. We believe this work helps to boost AI development and make it more accessible, efficient and trustworthy.
\end{abstract}

\section{Introduction}

Since the Transformer \cite{Vaswani2017-bg} was introduced in the Natural Language Processing (NLP) community, the advances in this field have increased exponentially~\cite{hf2020transformers}. 

Starting from popular models such as BERT~\cite{Devlin2018BERTPO} or Generative Pre-trained Transformers (GPT)~\citep{radford2018gpt} -both introduced in 2018-, researchers have been pushing the limits of scaling and learned representations in language models~\cite{Liu2019RoBERTaAR, radford2019gpt2, brown2020gpt3, chowdhery2022palm}.

Recent advances in Large Language Models (LLMs) are all over the news; these models represent a revolution in Artificial Intelligence (AI) due to their real-world applications through natural language processing (NLP), from internet chatbots to virtual assistants and programmers. However, these also pose significant risks and challenges. The most popular models (\eg chatGPT~\cite{openai2023gpt4}) are proprietary and not truly open-source, either transparent regarding their training data.

\begin{figure}[!t]
    \centering
    \includegraphics[width=\linewidth]{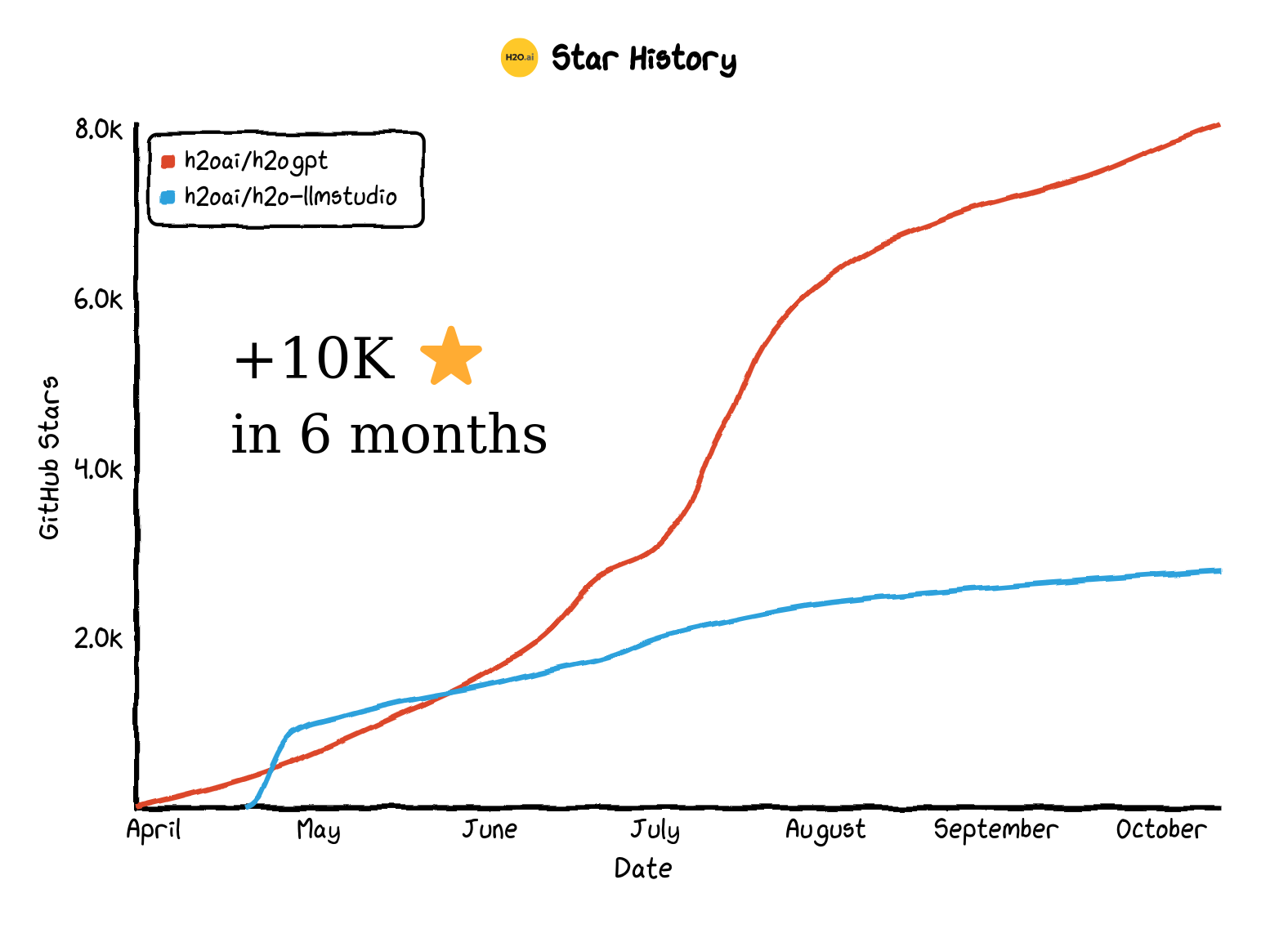}
    \vspace{-7mm}
    \caption{
    Evolution of our project in GitHub. Our tools have been widely adopted by the NLP community. See ~\url{https://github.com/h2oai/h2ogpt}.
    \vspace{-3mm}
    }
    \label{fig:h2ogpt-git}
\end{figure}

This fast advance leads to a wide range of practical challenges that must be addressed in order for these models to be widely utilized and explored.
The popularity and demand of LLMs call for systems to train, fine-tune, evaluate, scale, and deploy the models on a variety of platforms. Given the training costs (millions of dollars), practitioners increasingly rely on pre-trained general-purpose LLMs and fine-tune them for specific downstream tasks and datasets. This requires a wide catalogue of open-source pre-trained LLMs, and sophisticated procedures and tools for efficient fine-tuning. Moreover, considering the massive size of these models (usually from 7 to 100 Billion parameters), we also need compression techniques to deploy them successfully on different platforms.

We believe open-source language models help to boost AI development and make it more accessible and trustworthy. They lower entry hurdles, allowing people to tailor these models to their needs. This openness increases innovation, transparency, and fairness. As part of this effort, we \textbf{introduce two open-source libraries}: \textit{h2oGPT} and \textit{H2O LLM Studio}, for LLMs development, including Multi LLM deployment and evaluation --- widely adopted in the NLP community (see Fig.~\ref{fig:h2ogpt-git}).

\noindent\textbf{h2oGPT} (\url{https://github.com/h2oai/h2ogpt}) is a library dedicated to supporting open-source LLMs research, and facilitating their integration while ensuring privacy and transparency. Most integrated models are designed for both research and production. The main use-case of this library is to deploy and test efficiently a wide variety of LLMs on private databases and documents. This tool allows users to compare different models on several tasks and datasets concurrently. An example of this application is \url{https://gpt.h2o.ai/}.

\vspace{2mm}

\noindent\textbf{H2O LLM Studio} (\url{https://github.com/h2oai/h2o-llmstudio}) complements the previous library, and allows users to efficiently fine-tune any LLM using the most recent \sota techniques such as LoRA adapters~\cite{hu2021lora}, reinforcement learning (RLHF), and 4-bit training. After fine-tuning (or training), the models can be easily exported and deployed at the Hugging Face Hub~\footnote{\url{https://huggingface.co/models}}. Moreover, the library includes a graphic user interface (GUI) specially designed for large language models.

\vspace{2mm}

\textit{h2oGPT} and \textit{H2O LLM Studio} are an ongoing effort maintained frequently by the team of engineers and researchers at H2O.ai with exciting support from the open-source NLP community and external contributors. Both are released under the Apache 2.0 license~\footnote{\url{https://www.apache.org/licenses/LICENSE-2.0}}. Tutorials and detailed documentation are available at the corresponding websites and the technical report~\cite{candel2023h2ogpt}.

\section{Related Work}

Large language models (LLMs) are designed to process and understand vast amounts of natural language data \eg internet questions, text in documents, financial data, textbook material, etc. As foundation models~\citep{bommasani2021fundational}, these are trained from broad data at scale~\cite{Howard2018UniversalLM}, and can be adapted (\ie fine-tuned) to a wide range of down-stream tasks~\cite{Wang2018GLUEAM, Lewis2019-ak}. 

They are built on the \emph{Transformer} neural network architecture~\citep{Vaswani2017-bg}, which allows them to capture complex language patterns and relationships.
Derived from the Transformer, we find BERT-like models~\cite{devlin2018bert, le2020flaubert, Liu2019RoBERTaAR} focused on pre-training with bidirectional encoders. 
We also find the popular Generative Pre-trained Transformers (GPTs)~\cite{radford2018gpt, radford2019gpt2, brown2020gpt3, openai2023gpt4}, focused on generative pre-training. These serve as the engine of chatGPT.

Since 2022, we experience a new revolution in NLP with the rise of LLMs (over billion parameters models). These models usually follow a multi-stage training strategy, starting with a task-agnostic pre-training on large and diverse datasets. Some related LLMs are LLaMA~\cite{touvron2023llama}, GPT-NeoX~\cite{black2022gptneox20b}, BLOOM~\cite{scao2022bloom}, Palm~\cite{chowdhery2022palm}, OPT~\cite{zhang2022opt}, and GPT-4~\cite{openai2023gpt4}. We also explore community models such as Falcon~\cite{penedorefinedwebfalcon}, Alpaca~\cite{alpaca}, and OpenAssistant~\cite{kopf2023openassistant}.

\subsection{Why Open-Source LLMs?}

While commercially hosted and centralized LLMs like ChatGPT -based on GPT-4~\cite{openai2023gpt4}-, Microsoft's Bing AI Chat, and Google's Bard are powerful and effective, they have certain risks and limitations compared to open-source LLMs:

\begin{itemize}
    \itemsep0em 
    \item \textbf{Data Privacy and Security}: Many require sending data to external servers. This can raise concerns about data privacy, security, and compliance, especially for sensitive information or industries with strict regulations.
    \item \textbf{Dependency and Customization}: We want to allow users to train LLMs on private data safely, and customize the models to their specific needs and applications. Moreover the users can deploy them on their own infrastructure, and even modify the underlying code.
    \item \textbf{Traceability and Transparency}: To understand the risky behaviours of LLMs (\eg hallucinations, biases, private information etc.), and ensure their safe and trustworthy use, it is fundamental to analyze the dataset and training strategies used to produce such model.
    \item \textbf{Carbon footprint}: Users tend to adopt our open \sota models, instead of running expensive and complicated experiments (in most cases to replicate results). Therefore, we aim to reduce the overall carbon footprint (\ie GPU hours consumption) by providing high-quality models and tools.
\end{itemize}

Overall, open-source LLMs offer greater flexibility, control, and cost-effectiveness, while addressing data privacy and security concerns. 

\begin{figure*}[!t]
    \centering
    \includegraphics[width=\linewidth]{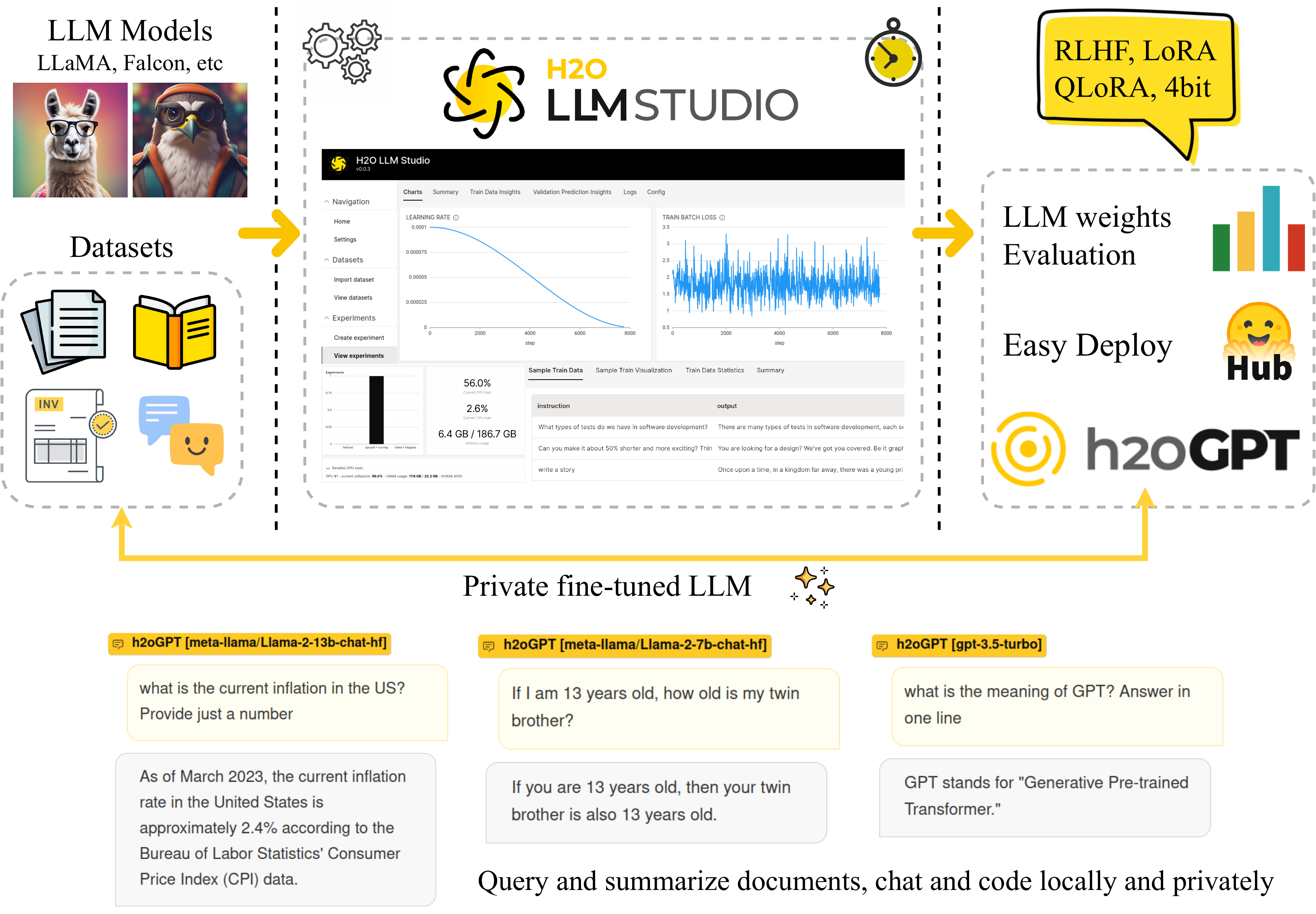}
    \caption{\textbf{Open LLM Ecosystem}. (left) The user does not need to transfer private data to 3rd parties, and can select any popular LLM \eg LLaMA, Falcon. (mid) H2O LLM Studio allows to train and fine-tune any language model using \sota techniques and a GUI without coding. (right) The models can be easily evaluated, exported and deployed. More information at \url{https://github.com/h2oai/h2o-llmstudio}. Apache 2 License.}
    \label{fig:teaser}
\end{figure*}

\section{H2O LLM Studio}
\label{sec:llmstudio}

\begin{figure*}[!t]
    \centering
    \includegraphics[width=\linewidth]{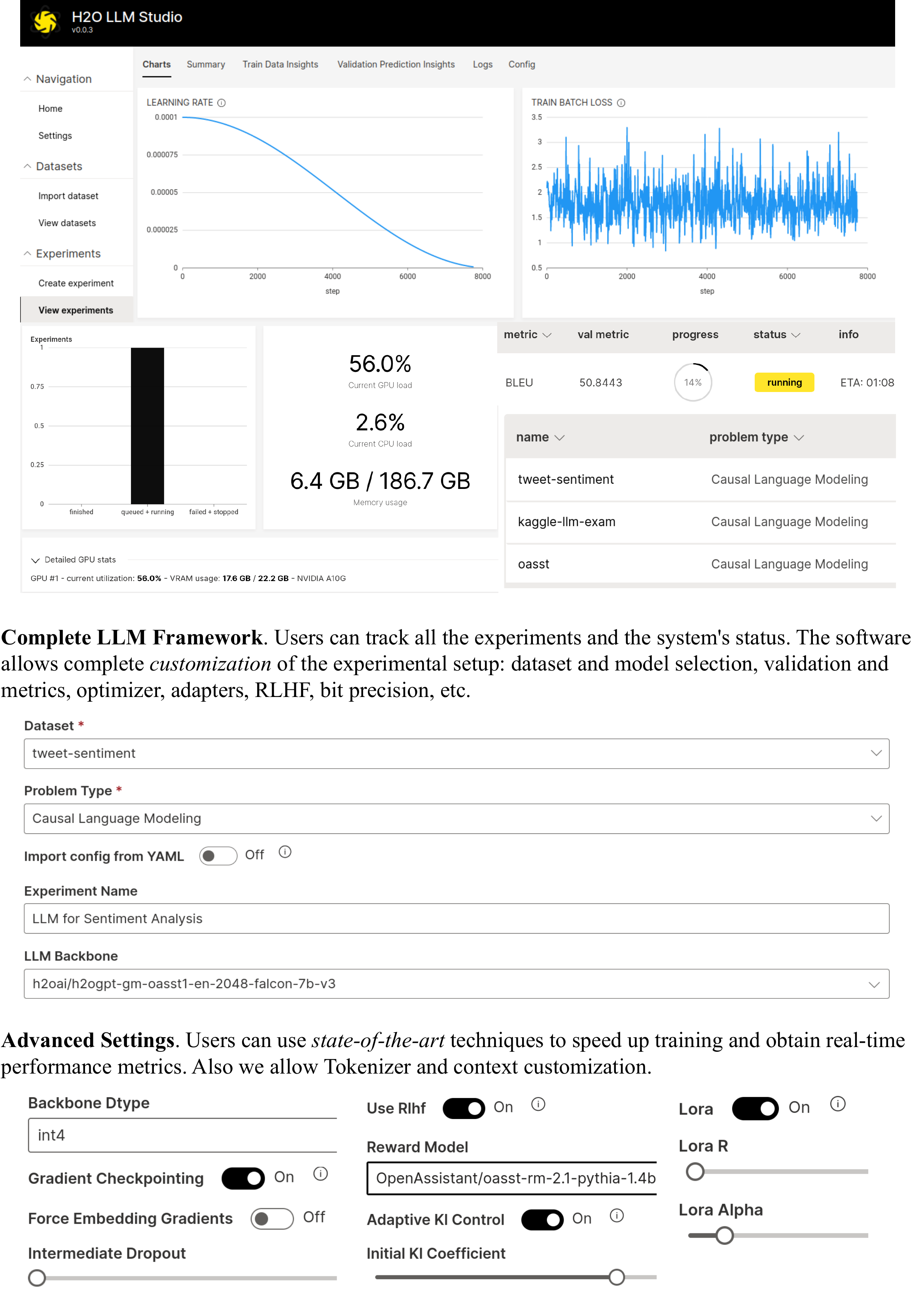}
    \caption{\textbf{LLM Studio} allows efficient training and fine-tuning of LLMs using \sota techniques (\eg advanced models, LoRA, int4, RLHF), and an intuitive GUI with complete experiment's customization. More information in \url{https://github.com/h2oai/h2o-llmstudio}. Apache 2 License.}
    \label{fig:llmstudio}
\end{figure*}

An open-source framework for efficient fine-tuning LLMs without coding, using a graphic user interface (GUI) specially designed for large language models~\footnote{\url{https://github.com/h2oai/h2o-llmstudio}}. This is illustrated in Figures~\ref{fig:teaser} and \ref{fig:llmstudio}.

\vspace{2mm}

We use the most popular \textbf{adapters} for fast fine-tuning such as Low-Rank Adaptation (LoRA)~\cite{hu2021lora} and QLoRA~\cite{dettmers2023qlora}, as well as 8-bit (up to 4-bit) model training with a low memory footprint, and the corresponding \textbf{quantization}. This allows to fine-tune small LLMs in regular GPUs, even using Google Colab or Kaggle. For example $<10$B models (\eg LlaMa-2 7B) can be fine-tuned in a single NVIDIA-T4 (16GB).

\vspace{2mm}

We also integrate \textit{Reinforcement Learning from Human Feedback (\textbf{RLHF})}~\cite{ouyang2022trainingrl, stiennon2020learningrl}. This feature is inspired in TRL~\footnote{\url{https://github.com/lvwerra/trl}}~\cite{vonwerra2022trl}, with the Proximal Policy Optimisation (PPO) by~\cite{ziegler2019fine}. 

LLM Studio allows complete \textbf{customization} of the experimental setup: dataset, \sota model selection, optimizer, learning rate schedule, tokenizer, sequence length (number of tokens), low-rank adapter, validation set and metrics, etc.

The users can \textbf{track} several simultaneous experiments, and easily \textbf{export} the logs and results. Moreover, the models can be easily exported to the Hugging Face Hub, to be shared with the community or deploy locally and privately.

The framework supports \textbf{any open-source language model}, we here highlight the most popular \sota large models: GPT-NeoX~\cite{black2022gptneox20b}, Falcon~\cite{penedorefinedwebfalcon}, LLaMa and Llama 2~\cite{touvron2023llama2}, Vicuna~\cite{vicuna2023}, WizardLM~\cite{xu2023wizardlm, luo2023wizardcoder}, h2oGPT~\cite{candel2023h2ogpt}, and MPT~\cite{mpt2023}. We summarize these models in Table~\ref{tab:models}. Most models are trained on a large amount of data (over 1T tokens), they can handle extremely long inputs (large context length), and are licensed for commercial use.

\begin{table}[]
    \centering
    \begin{tabular}{l c}
         \toprule
         Model & Size (B) \\
         \midrule
         Llama 2~\cite{touvron2023llama2} & 7 / 13 / 70 \\
         CodeLlama~\cite{touvron2023llama2} & 34 \\
         Falcon~\cite{penedorefinedwebfalcon} & 7 / 40 / 180 \\
         Mistral AI~\cite{mistral} & 7  \\
         GPT-NeoX~\cite{black2022gptneox20b} & 20 \\
         WizardLM~\cite{xu2023wizardlm} & 7 / 13 / 70 \\
         Vicuna~\cite{vicuna2023} & 13 \\
         MPT~\cite{mpt2023} & 7 / 30 \\
         h2oGPT~\cite{candel2023h2ogpt} & 7 to 70 \\
         GPT-3.5 (\emph{by OpenAI}) & ? \\
         \bottomrule
    \end{tabular}
    \caption{Most popular pre-trained LLMs for fine-tuning. We report the size in Billions (B) of parameters.
    \vspace{-2mm}
    }
    \label{tab:models}
\end{table}

We acknowledge \textbf{other existing tools} such as LLMTune~\cite{llmtune} and EasyLM~\cite{geng2023easylm}. However, these do not include as many features as LLM Studio (\eg GUI, supported models and techniques, etc), their licenses can be less permissive. Our tools are amongst the most adopted LLM-related software in GitHub (considering stars and forks by July 2023) --- see Fig.~\ref{fig:h2ogpt-git}.

\section{Multi LLM Deployment and Evaluation}
\label{sec:gpth2o}

Any model produced from LLM Studio can be easily integrated into HuggingFace's space \& models. We refer to our own space for more information and access to our models~\footnote{\url{https://huggingface.co/h2oai}}.

\begin{figure*}[!t]
    \centering
    \includegraphics[width=0.98\linewidth]{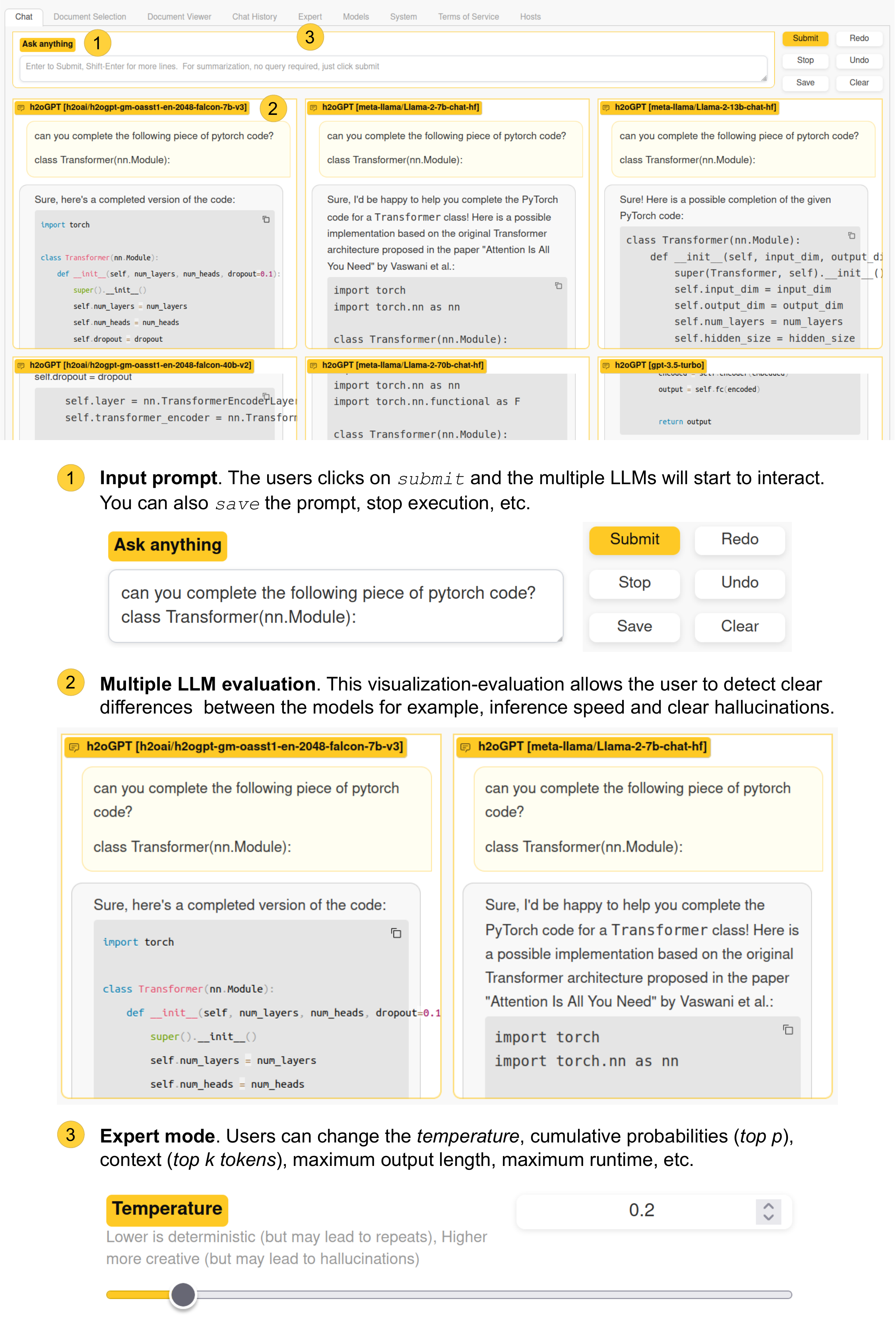}
    \caption{\textbf{h2oGPT}. Evaluation of multiple \emph{state-of-the-art} LLM models using the same prompt. This visualization and evaluation allows the user to detect clear differences between the models \emph{e.g.} faster or slower inference, clear hallucinations, common memorized patterns. Demo available at \url{https://gpt.h2o.ai/} completely free.}
    \label{fig:h2ogpt1}
\end{figure*}

In Fig.~\ref{fig:h2ogpt1} (top) we show a snapshot of our demo h2oGPT \url{https://gpt.h2o.ai/}. We deploy multiple \emph{state-of-the-art} LLM models including Falcon (7/40B), Llama 2 (7/13/70B), and GPT-3.5. This allows us to compare different models and setups.

\vspace{1mm}

The user's prompt is evaluated by the different LLMs \textbf{concurrently}. We can see the answer generation progress for each model, at the same time. Using this software we can identify clear differences between LLMs easily, for example fast/low inference, hallucinations, common response patterns, bias, memorized data etc. Also, we can analyze the effect of \textbf{prompt engineering} on the different models and expose vulnerabilities.
The users can deploy the models on a wide variety of inference servers (HF TGI server, vLLM, Gradio, OpenAI), and evaluate performance using reward models.

\paragraph{Document Analysis}
\textit{h2oGPT} also allows to query and summarize documents in many formats (\eg PDFs, Word, Code, Text, MarkDown, etc).
We implement an efficient use of context using instruct-tuned LLMs (no need for LangChain).

Note that this ecosystem can be reproduced locally, to analyze the models in a private  and safe manner.
We also provide a OpenAI-compliant Python client API for client-server control.

\vspace{3mm}

\paragraph{Guides \& Material} We provide a short \href{https://www.youtube.com/watch?v=_iktbj4obAI}{\textbf{Video tutorial (2 mins)}}, and a complete \href{https://youtu.be/Coj72EzmX20}{video overview} of the ecosystem (16 min, 340K views) on YouTube. 

Also a step-by-step tutorial \textbf{\href{https://www.youtube.com/watch?v=2orLCp984HA}{Make Your Own GPT With h2oGPT \& H2O LLM Studio}} (1hr).

We also host all of our models in HF: \url{https://huggingface.co/h2oai}. We refer the reader to our GitHubs for more demos, and documentation.

\section{Future Work}

Our open-source LLM Ecosystem is in constant development, \textit{h2oGPT} and \textit{LLM Studio} are updated based on the most recent research advances and demands. We plan to integrate new model quantization techniques, distillation and long-context training (context length over 100K tokens).

We also plan to support more multi-lingual models, and multi-modal models.

\section{Limitations}

\noindent\textbf{Datasets} Fine-tuning requires data text pairs of \texttt{instruction} and expected \texttt{result/answer}.

\noindent\textbf{Biases and Offensiveness} LLMs are trained on a diverse range of unfiltered internet text data, which may contain biased, racist, offensive, or otherwise inappropriate content. Therefore, the generated content by these models may sometimes exhibit biases or produce content that is offensive or inappropriate. We do not endorse, support, or promote any such content or viewpoints.

\noindent\textbf{Usage} The large language model is an AI-based tool and not a human. It may produce incorrect, offensive, nonsensical, or irrelevant responses. It is the user's responsibility to critically evaluate the generated content and use it at their discretion.

\noindent\textbf{Carbon footprint} Training LLMs is expensive and their use is associated to tons of CO$_{2}$ emissions~\cite{touvron2023llama}.

\noindent\textbf{Hallucinations} LLMs are probabilistic, therefore, certain ``random" behaviour is natural and expected, especially on complex prompts (\eg logical paradoxes, reasoning problems, etc) and ``unknown content" not present in the training corpus.

\section*{Broad Impact}

We advocate for the use of open-source LLMs to accelerate AI development and enhance its transparency, accessibility, security, and reliability. Our open framework for training, fine-tuning, deployment and analysis of LLMs enables this to any user, in a private and safe manner.
We provide a detailed \href{https://github.com/h2oai/h2ogpt#disclaimer}{\textbf{Disclaimer}} for users of our software, where we encourage users to use the LLMs responsibly and ethically.


\bibliography{refs}

\begin{thebibliography}{37}
\expandafter\ifx\csname natexlab\endcsname\relax\def\natexlab#1{#1}\fi

\bibitem[{Black et~al.(2022)Black, Biderman, Hallahan, Anthony, Gao, Golding, He, Leahy, McDonell, Phang et~al.}]{black2022gptneox20b}
Sid Black, Stella Biderman, Eric Hallahan, Quentin Anthony, Leo Gao, Laurence Golding, Horace He, Connor Leahy, Kyle McDonell, Jason Phang, et~al. 2022.
\newblock Gpt-neox-20b: An open-source autoregressive language model.
\newblock \emph{arXiv preprint arXiv:2204.06745}.

\bibitem[{Bommasani et~al.(2021)Bommasani, Hudson, Adeli, Altman, Arora, von Arx, Bernstein, Bohg, Bosselut, Brunskill et~al.}]{bommasani2021fundational}
Rishi Bommasani, Drew~A Hudson, Ehsan Adeli, Russ Altman, Simran Arora, Sydney von Arx, Michael~S Bernstein, Jeannette Bohg, Antoine Bosselut, Emma Brunskill, et~al. 2021.
\newblock On the opportunities and risks of foundation models.
\newblock \emph{arXiv preprint arXiv:2108.07258}.

\bibitem[{Brown et~al.(2020)Brown, Mann, Ryder, Subbiah, Kaplan, Dhariwal, Neelakantan, Shyam, Sastry, Askell, Agarwal, Herbert-Voss, Krueger, Henighan, Child, Ramesh, Ziegler, Wu, Winter, Hesse, Chen, Sigler, Litwin, Gray, Chess, Clark, Berner, McCandlish, Radford, Sutskever, and Amodei}]{brown2020gpt3}
Tom~B. Brown, Benjamin Mann, Nick Ryder, Melanie Subbiah, Jared Kaplan, Prafulla Dhariwal, Arvind Neelakantan, Pranav Shyam, Girish Sastry, Amanda Askell, Sandhini Agarwal, Ariel Herbert-Voss, Gretchen Krueger, Tom Henighan, Rewon Child, Aditya Ramesh, Daniel~M. Ziegler, Jeffrey Wu, Clemens Winter, Christopher Hesse, Mark Chen, Eric Sigler, Mateusz Litwin, Scott Gray, Benjamin Chess, Jack Clark, Christopher Berner, Sam McCandlish, Alec Radford, Ilya Sutskever, and Dario Amodei. 2020.
\newblock \href {https://doi.org/10.48550/ARXIV.2005.14165} {Language models are few-shot learners}.

\bibitem[{Candel et~al.(2023)Candel, McKinney, Singer, Pfeiffer, Jeblick, Prabhu, Gambera, Landry, Bansal, Chesler et~al.}]{candel2023h2ogpt}
Arno Candel, Jon McKinney, Philipp Singer, Pascal Pfeiffer, Maximilian Jeblick, Prithvi Prabhu, Jeff Gambera, Mark Landry, Shivam Bansal, Ryan Chesler, et~al. 2023.
\newblock h2ogpt: Democratizing large language models.
\newblock \emph{arXiv preprint arXiv:2306.08161}.

\bibitem[{Chiang et~al.(2023)Chiang, Li, Lin, Sheng, Wu, Zhang, Zheng, Zhuang, Zhuang, Gonzalez, Stoica, and Xing}]{vicuna2023}
Wei-Lin Chiang, Zhuohan Li, Zi~Lin, Ying Sheng, Zhanghao Wu, Hao Zhang, Lianmin Zheng, Siyuan Zhuang, Yonghao Zhuang, Joseph~E. Gonzalez, Ion Stoica, and Eric~P. Xing. 2023.
\newblock \href {https://lmsys.org/blog/2023-03-30-vicuna/} {Vicuna: An open-source chatbot impressing gpt-4 with 90\%* chatgpt quality}.

\bibitem[{Chowdhery et~al.(2022)Chowdhery, Narang, Devlin, Bosma, Mishra, Roberts, Barham, Chung, Sutton, Gehrmann, Schuh, Shi, Tsvyashchenko, Maynez, Rao, Barnes, Tay, Shazeer, Prabhakaran, Reif, Du, Hutchinson, Pope, Bradbury, Austin, Isard, Gur-Ari, Yin, Duke, Levskaya, Ghemawat, Dev, Michalewski, Garcia, Misra, Robinson, Fedus, Zhou, Ippolito, Luan, Lim, Zoph, Spiridonov, Sepassi, Dohan, Agrawal, Omernick, Dai, Pillai, Pellat, Lewkowycz, Moreira, Child, Polozov, Lee, Zhou, Wang, Saeta, Diaz, Firat, Catasta, Wei, Meier-Hellstern, Eck, Dean, Petrov, and Fiedel}]{chowdhery2022palm}
Aakanksha Chowdhery, Sharan Narang, Jacob Devlin, Maarten Bosma, Gaurav Mishra, Adam Roberts, Paul Barham, Hyung~Won Chung, Charles Sutton, Sebastian Gehrmann, Parker Schuh, Kensen Shi, Sasha Tsvyashchenko, Joshua Maynez, Abhishek Rao, Parker Barnes, Yi~Tay, Noam Shazeer, Vinodkumar Prabhakaran, Emily Reif, Nan Du, Ben Hutchinson, Reiner Pope, James Bradbury, Jacob Austin, Michael Isard, Guy Gur-Ari, Pengcheng Yin, Toju Duke, Anselm Levskaya, Sanjay Ghemawat, Sunipa Dev, Henryk Michalewski, Xavier Garcia, Vedant Misra, Kevin Robinson, Liam Fedus, Denny Zhou, Daphne Ippolito, David Luan, Hyeontaek Lim, Barret Zoph, Alexander Spiridonov, Ryan Sepassi, David Dohan, Shivani Agrawal, Mark Omernick, Andrew~M. Dai, Thanumalayan~Sankaranarayana Pillai, Marie Pellat, Aitor Lewkowycz, Erica Moreira, Rewon Child, Oleksandr Polozov, Katherine Lee, Zongwei Zhou, Xuezhi Wang, Brennan Saeta, Mark Diaz, Orhan Firat, Michele Catasta, Jason Wei, Kathy Meier-Hellstern, Douglas Eck, Jeff Dean, Slav Petrov, and Noah Fiedel. 2022.
\newblock \href {https://arxiv.org/abs/2204.02311} {Palm: Scaling language modeling with pathways}.

\bibitem[{Dettmers et~al.(2023)Dettmers, Pagnoni, Holtzman, and Zettlemoyer}]{dettmers2023qlora}
Tim Dettmers, Artidoro Pagnoni, Ari Holtzman, and Luke Zettlemoyer. 2023.
\newblock Qlora: Efficient finetuning of quantized llms.
\newblock \emph{arXiv preprint arXiv:2305.14314}.

\bibitem[{Devlin et~al.(2018{\natexlab{a}})Devlin, Chang, Lee, and Toutanova}]{Devlin2018BERTPO}
Jacob Devlin, Ming-Wei Chang, Kenton Lee, and Kristina Toutanova. 2018{\natexlab{a}}.
\newblock Bert: Pre-training of deep bidirectional transformers for language understanding.
\newblock In \emph{NAACL-HLT}.

\bibitem[{Devlin et~al.(2018{\natexlab{b}})Devlin, Chang, Lee, and Toutanova}]{devlin2018bert}
Jacob Devlin, Ming-Wei Chang, Kenton Lee, and Kristina Toutanova. 2018{\natexlab{b}}.
\newblock Bert: Pre-training of deep bidirectional transformers for language understanding.
\newblock \emph{arXiv preprint arXiv:1810.04805}.

\bibitem[{Geng(2023)}]{geng2023easylm}
Xinyang Geng. 2023.
\newblock \href {https://github.com/young-geng/EasyLM} {Easylm: A simple and scalable training framework for large language models}.

\bibitem[{Howard and Ruder(2018)}]{Howard2018UniversalLM}
Jeremy Howard and Sebastian Ruder. 2018.
\newblock Universal language model fine-tuning for text classification.
\newblock In \emph{ACL}.

\bibitem[{Hu et~al.(2021)Hu, Shen, Wallis, Allen-Zhu, Li, Wang, Wang, and Chen}]{hu2021lora}
Edward~J Hu, Yelong Shen, Phillip Wallis, Zeyuan Allen-Zhu, Yuanzhi Li, Shean Wang, Lu~Wang, and Weizhu Chen. 2021.
\newblock Lora: Low-rank adaptation of large language models.
\newblock \emph{arXiv preprint arXiv:2106.09685}.

\bibitem[{K{\"o}pf et~al.(2023)K{\"o}pf, Kilcher, von R{\"u}tte, Anagnostidis, Tam, Stevens, Barhoum, Duc, Stanley, Nagyfi et~al.}]{kopf2023openassistant}
Andreas K{\"o}pf, Yannic Kilcher, Dimitri von R{\"u}tte, Sotiris Anagnostidis, Zhi-Rui Tam, Keith Stevens, Abdullah Barhoum, Nguyen~Minh Duc, Oliver Stanley, Rich{\'a}rd Nagyfi, et~al. 2023.
\newblock Openassistant conversations--democratizing large language model alignment.
\newblock \emph{arXiv preprint arXiv:2304.07327}.

\bibitem[{Kuleshov(2023)}]{llmtune}
Volodymyr Kuleshov. 2023.
\newblock Llmtune: Fine-tuning large language models on one consumer gpu.
\newblock \url{https://github.com/kuleshov-group/llmtune}.

\bibitem[{Le et~al.(2020)Le, Vial, Frej, Segonne, Coavoux, Lecouteux, Allauzen, Crabb\'{e}, Besacier, and Schwab}]{le2020flaubert}
Hang Le, Lo\"{i}c Vial, Jibril Frej, Vincent Segonne, Maximin Coavoux, Benjamin Lecouteux, Alexandre Allauzen, Beno\^{i}t Crabb\'{e}, Laurent Besacier, and Didier Schwab. 2020.
\newblock \href {https://www.aclweb.org/anthology/2020.lrec-1.302} {Flaubert: Unsupervised language model pre-training for french}.
\newblock In \emph{Proceedings of The 12th Language Resources and Evaluation Conference}, pages 2479--2490, Marseille, France. European Language Resources Association.

\bibitem[{Lewis et~al.(2019)Lewis, Liu, Goyal, Ghazvininejad, Mohamed, Levy, Stoyanov, and Zettlemoyer}]{Lewis2019-ak}
Mike Lewis, Yinhan Liu, Naman Goyal, Marjan Ghazvininejad, Abdelrahman Mohamed, Omer Levy, Ves Stoyanov, and Luke Zettlemoyer. 2019.
\newblock \href {http://arxiv.org/abs/1910.13461} {{BART}: Denoising {Sequence-to-Sequence} pre-training for natural language generation, translation, and comprehension}.

\bibitem[{Liu et~al.(2019)Liu, Ott, Goyal, Du, Joshi, Chen, Levy, Lewis, Zettlemoyer, and Stoyanov}]{Liu2019RoBERTaAR}
Yinhan Liu, Myle Ott, Naman Goyal, Jingfei Du, Mandar~S. Joshi, Danqi Chen, Omer Levy, Mike Lewis, Luke~S. Zettlemoyer, and Veselin Stoyanov. 2019.
\newblock Roberta: A robustly optimized bert pretraining approach.
\newblock \emph{ArXiv}, abs/1907.11692.

\bibitem[{Luo et~al.(2023)Luo, Xu, Zhao, Sun, Geng, Hu, Tao, Ma, Lin, and Jiang}]{luo2023wizardcoder}
Ziyang Luo, Can Xu, Pu~Zhao, Qingfeng Sun, Xiubo Geng, Wenxiang Hu, Chongyang Tao, Jing Ma, Qingwei Lin, and Daxin Jiang. 2023.
\newblock \href {http://arxiv.org/abs/2306.08568} {Wizardcoder: Empowering code large language models with evol-instruct}.

\bibitem[{{Mistral AI}(2023)}]{mistral}
{Mistral AI}. 2023.
\newblock Mistral 7b introduction.
\newblock \url{https://mistral.ai/news/announcing-mistral-7b/}.

\bibitem[{MosaicML(2023)}]{mpt2023}
MosaicML. 2023.
\newblock \href {https://www.mosaicml.com/blog/mpt-30b} {Mpt-30b: Raising the bar for open-source foundation models}.

\bibitem[{OpenAI(2023)}]{openai2023gpt4}
OpenAI. 2023.
\newblock \href {http://arxiv.org/abs/2303.08774} {Gpt-4 technical report}.

\bibitem[{Ouyang et~al.(2022)Ouyang, Wu, Jiang, Almeida, Wainwright, Mishkin, Zhang, Agarwal, Slama, Ray et~al.}]{ouyang2022trainingrl}
Long Ouyang, Jeffrey Wu, Xu~Jiang, Diogo Almeida, Carroll Wainwright, Pamela Mishkin, Chong Zhang, Sandhini Agarwal, Katarina Slama, Alex Ray, et~al. 2022.
\newblock Training language models to follow instructions with human feedback.
\newblock \emph{Advances in Neural Information Processing Systems}, 35:27730--27744.

\bibitem[{Penedo et~al.()Penedo, Malartic, Hesslow, Cojocaru, Cappelli, Pannier, Almazrouei, and Launay}]{penedorefinedwebfalcon}
Guilherme Penedo, Quentin Malartic, Daniel Hesslow, Ruxandra Cojocaru, Alessandro Cappelli, Baptiste Pannier, Ebtesam Almazrouei, and Julien Launay.
\newblock The refinedweb dataset for falcon llm: Outperforming curated corpora with web data, and web data only.

\bibitem[{Radford et~al.(2018)Radford, Narasimhan, Salimans, Sutskever et~al.}]{radford2018gpt}
Alec Radford, Karthik Narasimhan, Tim Salimans, Ilya Sutskever, et~al. 2018.
\newblock Improving language understanding by generative pre-training.

\bibitem[{Radford et~al.(2019)Radford, Wu, Child, Luan, Amodei, Sutskever et~al.}]{radford2019gpt2}
Alec Radford, Jeffrey Wu, Rewon Child, David Luan, Dario Amodei, Ilya Sutskever, et~al. 2019.
\newblock Language models are unsupervised multitask learners.
\newblock \emph{OpenAI blog}, 1(8):9.

\bibitem[{Scao et~al.(2022)Scao, Fan, Akiki, Pavlick, Ili{\'c}, Hesslow, Castagn{\'e}, Luccioni, Yvon, Gall{\'e} et~al.}]{scao2022bloom}
Teven~Le Scao, Angela Fan, Christopher Akiki, Ellie Pavlick, Suzana Ili{\'c}, Daniel Hesslow, Roman Castagn{\'e}, Alexandra~Sasha Luccioni, Fran{\c{c}}ois Yvon, Matthias Gall{\'e}, et~al. 2022.
\newblock Bloom: A 176b-parameter open-access multilingual language model.
\newblock \emph{arXiv preprint arXiv:2211.05100}.

\bibitem[{Stiennon et~al.(2020)Stiennon, Ouyang, Wu, Ziegler, Lowe, Voss, Radford, Amodei, and Christiano}]{stiennon2020learningrl}
Nisan Stiennon, Long Ouyang, Jeffrey Wu, Daniel Ziegler, Ryan Lowe, Chelsea Voss, Alec Radford, Dario Amodei, and Paul~F Christiano. 2020.
\newblock Learning to summarize with human feedback.
\newblock \emph{Advances in Neural Information Processing Systems}, 33:3008--3021.

\bibitem[{Taori et~al.(2023)Taori, Gulrajani, Zhang, Dubois, Li, Guestrin, Liang, and Hashimoto}]{alpaca}
Rohan Taori, Ishaan Gulrajani, Tianyi Zhang, Yann Dubois, Xuechen Li, Carlos Guestrin, Percy Liang, and Tatsunori~B. Hashimoto. 2023.
\newblock Stanford alpaca: An instruction-following llama model.
\newblock \url{https://github.com/tatsu-lab/stanford_alpaca}.

\bibitem[{Touvron et~al.(2023{\natexlab{a}})Touvron, Lavril, Izacard, Martinet, Lachaux, Lacroix, Rozi{\`e}re, Goyal, Hambro, Azhar et~al.}]{touvron2023llama}
Hugo Touvron, Thibaut Lavril, Gautier Izacard, Xavier Martinet, Marie-Anne Lachaux, Timoth{\'e}e Lacroix, Baptiste Rozi{\`e}re, Naman Goyal, Eric Hambro, Faisal Azhar, et~al. 2023{\natexlab{a}}.
\newblock Llama: Open and efficient foundation language models.
\newblock \emph{arXiv preprint arXiv:2302.13971}.

\bibitem[{Touvron et~al.(2023{\natexlab{b}})Touvron, Martin, Stone, Albert, Almahairi, Babaei, Bashlykov, Batra, Bhargava, Bhosale et~al.}]{touvron2023llama2}
Hugo Touvron, Louis Martin, Kevin Stone, Peter Albert, Amjad Almahairi, Yasmine Babaei, Nikolay Bashlykov, Soumya Batra, Prajjwal Bhargava, Shruti Bhosale, et~al. 2023{\natexlab{b}}.
\newblock Llama 2: Open foundation and fine-tuned chat models.
\newblock \emph{arXiv preprint arXiv:2307.09288}.

\bibitem[{Vaswani et~al.(2017)Vaswani, Shazeer, Parmar, Uszkoreit, Jones, Gomez, Kaiser, and Polosukhin}]{Vaswani2017-bg}
Ashish Vaswani, Noam Shazeer, Niki Parmar, Jakob Uszkoreit, Llion Jones, Aidan~N Gomez, {\L}~Ukasz Kaiser, and Illia Polosukhin. 2017.
\newblock Attention is all you need.
\newblock In I~Guyon, U~V Luxburg, S~Bengio, H~Wallach, R~Fergus, S~Vishwanathan, and R~Garnett, editors, \emph{Advances in Neural Information Processing Systems 30}, pages 5998--6008. Curran Associates, Inc.

\bibitem[{von Werra et~al.(2020)von Werra, Belkada, Tunstall, Beeching, Thrush, and Lambert}]{vonwerra2022trl}
Leandro von Werra, Younes Belkada, Lewis Tunstall, Edward Beeching, Tristan Thrush, and Nathan Lambert. 2020.
\newblock Trl: Transformer reinforcement learning.
\newblock \url{https://github.com/lvwerra/trl}.

\bibitem[{Wang et~al.(2018)Wang, Singh, Michael, Hill, Levy, and Bowman}]{Wang2018GLUEAM}
Alex Wang, Amanpreet Singh, Julian Michael, Felix Hill, Omer Levy, and Samuel~R. Bowman. 2018.
\newblock Glue: A multi-task benchmark and analysis platform for natural language understanding.
\newblock In \emph{ICLR}.

\bibitem[{Wolf et~al.(2020)Wolf, Debut, Sanh, Chaumond, Delangue, Moi, Cistac, Rault, Louf, Funtowicz et~al.}]{hf2020transformers}
Thomas Wolf, Lysandre Debut, Victor Sanh, Julien Chaumond, Clement Delangue, Anthony Moi, Pierric Cistac, Tim Rault, R{\'e}mi Louf, Morgan Funtowicz, et~al. 2020.
\newblock Transformers: State-of-the-art natural language processing.
\newblock In \emph{Proceedings of the 2020 conference on empirical methods in natural language processing: system demonstrations}, pages 38--45.

\bibitem[{Xu et~al.(2023)Xu, Sun, Zheng, Geng, Zhao, Feng, Tao, and Jiang}]{xu2023wizardlm}
Can Xu, Qingfeng Sun, Kai Zheng, Xiubo Geng, Pu~Zhao, Jiazhan Feng, Chongyang Tao, and Daxin Jiang. 2023.
\newblock Wizardlm: Empowering large language models to follow complex instructions.
\newblock \emph{arXiv preprint arXiv:2304.12244}.

\bibitem[{Zhang et~al.(2022)Zhang, Roller, Goyal, Artetxe, Chen, Chen, Dewan, Diab, Li, Lin et~al.}]{zhang2022opt}
Susan Zhang, Stephen Roller, Naman Goyal, Mikel Artetxe, Moya Chen, Shuohui Chen, Christopher Dewan, Mona Diab, Xian Li, Xi~Victoria Lin, et~al. 2022.
\newblock Opt: Open pre-trained transformer language models.
\newblock \emph{arXiv preprint arXiv:2205.01068}.

\bibitem[{Ziegler et~al.(2019)Ziegler, Stiennon, Wu, Brown, Radford, Amodei, Christiano, and Irving}]{ziegler2019fine}
Daniel~M Ziegler, Nisan Stiennon, Jeffrey Wu, Tom~B Brown, Alec Radford, Dario Amodei, Paul Christiano, and Geoffrey Irving. 2019.
\newblock Fine-tuning language models from human preferences.
\newblock \emph{arXiv preprint arXiv:1909.08593}.

\end{thebibliography}
\bibliographystyle{acl_natbib}






\end{document}